\documentclass{ws-procs11x85}
\usepackage{ws-procs-thm}           
\usepackage{booktabs}
\usepackage{multirow}
\usepackage{hyperref}

\begin{document}
\title{Self-omics: A Self-supervised Learning Framework for Multi-omics Cancer Data}

\author{Sayed Hashim$^\dag$, Karthik Nandakumar and Mohammad Yaqub}

\address{Mohamed Bin Zayed University of Artificial Intelligence\\
Abu Dhabi, UAE\\
$^\dag$E-mail: sayed.hashim@mbzuai.ac.ae\\
}



\begin{abstract}
We have gained access to vast amounts of multi-omics data thanks to Next Generation Sequencing. However, it is challenging to analyse this data due to its high dimensionality and much of it not being annotated. Lack of annotated data is a significant problem in machine learning, and Self-Supervised Learning (SSL) methods are typically used to deal with limited labelled data. However, there is a lack of studies that use SSL methods to exploit inter-omics relationships on unlabelled multi-omics data. In this work, we develop a novel and efficient pre-training paradigm that consists of various SSL components, including but not limited to contrastive alignment, data recovery from corrupted samples, and using one type of omics data to recover other omic types. Our pre-training paradigm improves performance on downstream tasks with limited labelled data. We show that our approach outperforms the state-of-the-art method in cancer type classification on the TCGA pan-cancer dataset in semi-supervised setting. Moreover, we show that the encoders that are pre-trained using our approach can be used as powerful feature extractors even without fine-tuning. Our ablation study shows that the method is not overly dependent on any pretext task component. The network architectures in our approach are designed to handle missing omic types and multiple datasets for pre-training and downstream training. Our pre-training paradigm can be extended to perform zero-shot classification of rare cancers. 
\end{abstract}


\copyrightinfo{\copyright\ 2022 The Authors. Open Access chapter published by World Scientific Publishing Company and distributed under the terms of the Creative Commons Attribution Non-Commercial (CC BY-NC) 4.0 License.}

\section{Introduction}

According to WHO, cancer accounted for around 10 million deaths in 2020 or about one in six deaths \cite{cancerwho}. 
Many cancers can be cured with early diagnosis, and effective treatment \cite{cancerearlydiag}. 
Various factors are responsible for late diagnoses, such as symptoms being detected late, lack of access to oncologists, as well as the time \& cost involved. It could also be because of vague and unclear symptoms and indistinguishable signs on scans and mammograms \cite{late_diag_reasons}. Nevertheless, performing cancer diagnosis in its early stages or even before it starts developing could remarkably improve survival and provide opportunities for more effective treatment.
Studies in the areas of biology that end with omics, such as genomics, proteomics, transcriptomics or metabolomics, are called omics sciences. With the advent of Next Generation Sequencing, we have gained access to multiple types of omics data. Each type of omics data reveals different characteristics within the tumour. However, due to the high dimensionality and the numerous different types of omics data, it is nearly impossible for clinicians to analyse multi-omics data. Due to this reason, they tend to focus on analysing the values of specific biomarkers. However, to get a complete picture of a tumour, which is heterogeneous and complex, multi-omics data analysis is vital.

Modern machine learning algorithms, especially deep neural networks, have shown to be able to work well with high-dimensional data. Deep learning has made massive progress in tasks like object recognition, object detection and semantic segmentation in the visual domain. It has also made strides in speech and natural language processing on tasks such as machine translation, speech recognition and question answering. The algorithms developed for the tasks mentioned above require processing high-dimensional inputs. 
In this work, we developed Self-Supervised Learning (SSL) methods for multi-omics data to provide supervision to the model from unlabelled data. 
We explored various SSL pretext tasks on top of the usual reconstruction task with autoencoders. Some of the SSL techniques we implemented include contrastive learning, recovering data from its corrupted versions and aligning representations from multi-omics data. 

The low-dimensional representations that our model produces from high-dimensional multi-omics data can be considered "computational biomarkers". The model that learns from large datasets gets good at producing such biomarkers and can be used to produce good representations for smaller datasets. Furthermore, as the model learns from tumours diagnosed early, it produces better representations for such tumours. Therefore, even if the dataset at hand does not have samples of tumours that are sequenced early, the fact that it was pre-trained on a large dataset that contains many samples of such tumours makes the model better at early diagnosis. 

\section{Literature Review}
Self-supervised learning (SSL) has been extensively applied in representation learning of data in various domains such as natural language processing \cite{ssl_nlp1, ssl_nlp2, ssl_nlp3} audio and image \cite{ssl_im1, ssl_im2, ssl_im3}. These methods mainly use spatial, semantic and temporal structural relationships in the data. This is done through developing novel pretext tasks, data augmentation methods and model architectures. Due to the absence of the relationships mentioned above in tabular data, such methods could be less effective. For instance, augmentation methods used on images, such as scaling and rotation, cannot be directly used on tabular data. SSL techniques have not been explored enough on tabular data due to these reasons \cite{vime}. 

An autoencoder is a deep network that consists of an encoder and decoder \cite{ae_bengio}. While the encoder is trained to map the input to a latent representation, the decoder is trained to reconstruct the input from this latent representation. A popular work in images is denoising autoencoders (DAE) \cite{denoisae}. It is built on the hypothesis that partially destroyed inputs should result in a similar latent representation as the original inputs. In this work, the authors investigated an autoencoder's robustness to partial demolition of inputs. The input is corrupted and fed to the autoencoder, whose job is to recover the original "clean" input. A group of researchers developed VIME \cite{vime}, a novel SSL framework for tabular data. They developed a couple of pretext tasks called feature vector estimation and mask vector estimation. The former aims to reconstruct an input sample from its masked version, while the latter involves predicting the mask vector applied to the sample. In other words, the pretext task is to estimate which features are masked and predict the values of the corrupted features. A work called SubTab focuses on converting the representation learning problem from single-view to multi-view \cite{subtab}. Here, the features are divided into subsets to produce the various views. The authors claim that this is analogous to cropping images and bagging features in ensemble learning. They demonstrate that the encoder learns more useful representations from a subset of the data than a corrupted version of it. They pre-trained the network on this pretext task and tested its performance on some downstream tasks.

Self-supervised representation learning of multi-omics data is an under-studied area of research. Many methods used for representation learning mainly focus on the integration of multi-omics data. Many integration strategies have been proposed. We will review the integration methods here due to the lack of self-supervised methods. A group developed a group lasso regularised deep learning method for cancer prognosis by integrating multi-omics data using early fusion \cite{early_1}. They perform various data preprocessing techniques, and the model consists of a few fully connected layers. Another work integrates multi-omics data using standard and disjointed deep autoencoders \cite{early_2}. Various omics data such as DNA methylation, microRNA expression, mRNA expression and reverse phase protein array data are concatenated before being fed into the autoencoder. A work called OmiEmbed \cite{omiembed} does intermediate multi-omics data integration. It is a multi-task framework that is built on a variational autoencoder. The pretext task here is the reconstruction of three types of omics data: gene expression, microRNA expression and DNA methylation. They show the effectiveness of their method by testing on various downstream tasks. They also developed a multi-task strategy that concurrently trains multiple downstream modules such as survival analysis, cancer type classification and phenotype prediction. Training it this way has shown to perform better than training the downstream modules separately. Late integration of multi-omics data was done in a work that predicts breast cancer prognosis \cite{late_1}. They perform feature selection and use a deep neural network for the task. Gene expression, copy number alterations and clinical information are fed into three separate networks. Their predictions are combined at the end with a score-fusing technique called weighted linear aggregation.

There exists a lack of studies on self-supervised representation learning of multi-omics data. Studies focusing on adding more pretext tasks on top of the reconstruction task are rare. The usual focus is on integrating the data and less on exploiting inter-omics relationships through constraints and other SSL losses. Moreover, lack of annotated data can be tackled with SSL approaches. 

\section{Method}
\subsection{Dataset}
For our experiments, we used The Cancer Genome Atlas (TCGA) pan-cancer multi-omics dataset \cite{17}. Table \ref{tab:tcga} gives an overview of the dataset. It is one of the most popular multi-omics datasets. It consists of omics data as well as phenotypic information of patients. We used three types of omics data from the TCGA dataset: DNA methylation, miRNA stem-loop expression, and gene expression. They are 485,577, 1881 and 60,483 dimensional respectively. The dataset contains samples of 33 different tumour types and of normal tissues.

\begin{table}[h]
\tbl{An overview of TCGA pan-cancer dataset.}
{\begin{tabular}{@{}lccc@{}}
\toprule
\textbf{Dataset} & \multicolumn{3}{c}{\textbf{TCGA}}                    \\ \midrule
Domain           & \multicolumn{3}{c}{Pan-cancer}                       \\
Tumour types     & \multicolumn{3}{c}{33 + 1(normal) = 34}              \\
Omics data type  & Gene exp & DNA methylation & miRNA exp \\
No of features   & 60,483          & 485,577         & 1881             \\
No of samples    & 11,538          & 9736            & 11,020           \\ \bottomrule
\\
\end{tabular}}
\label{tab:tcga}
\end{table}

\subsection{Data Preprocessing}
We downloaded harmonised data of 3 types of omics data from \href{https://xenabrowser.net/datapages/}{UCSC Xena data portal} \cite{ucsc}. RNA-Seq gene expression dataset comprises 60,483 features, each denoting the expression of a gene. Gene expression level is obtained as the $\log_2$ transformation of fragments per kilobase of transcript per million mapped reads (FPKM) value. miRNA stem-loop expression levels were given as the $\log_2$ transformation of reads per million mapped reads (RPM) value. DNA methylation dataset comprises beta values for each CpG site. Beta values are the ratio of methylated to total array intensity for the corresponding CpG site \cite{gdc}. Lower beta values mean lower levels of methylation and vice-versa. The beta values missing in the DNA methylation dataset were mean imputed. We removed the means of the three datasets and scaled them to unit variance.


\subsection{Pretext problem formulation}
\begin{figure}
    \centerline{
    \includegraphics[scale=0.36]{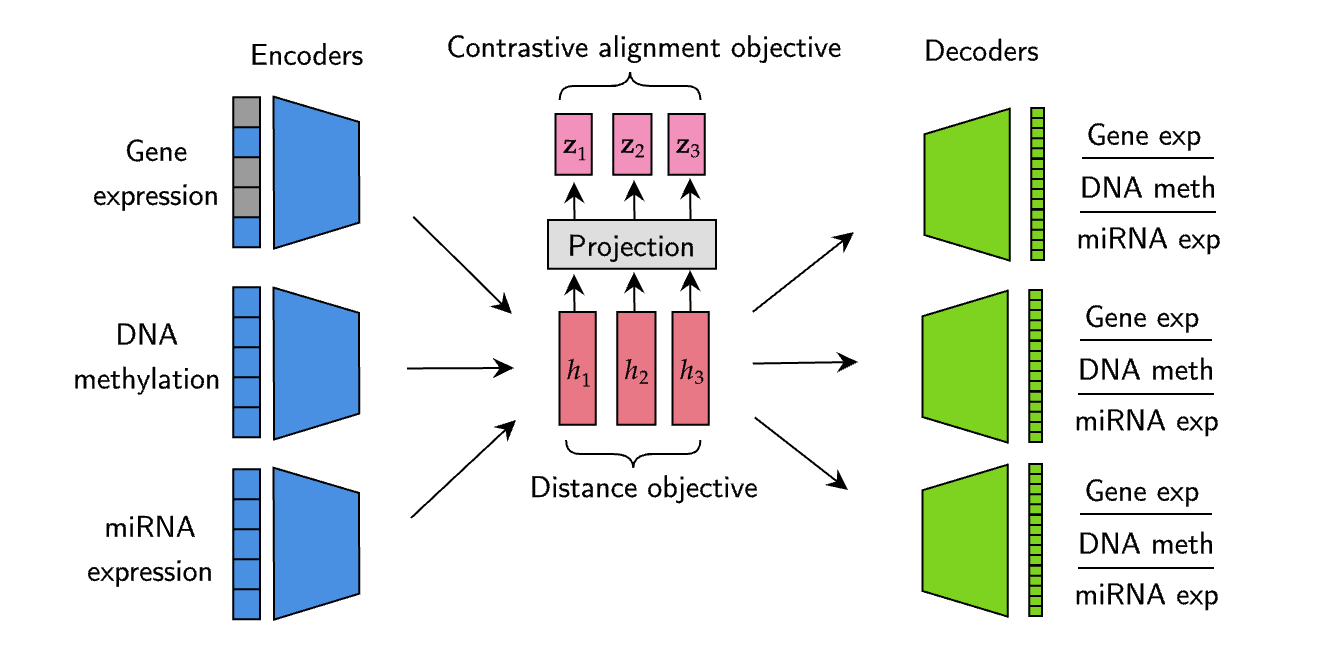}}
    \caption{Pre-training architecture: We partially masked the inputs and fed them to the encoders one by one, producing three latent representations, $h_1$, $h_2$ and $h_3$. These are then passed onto the decoders to reconstruct the entire feature set, including the other two omic types. The latent representations are also fed into a projection layer to compute contrastive alignment objective between each pair of projections, namely $(z_1,z_2)$, $(z_1,z_3)$ and $(z_2,z_3)$. Distance objectives between each pair of latent representations are also minimised. Pre-training also minimises contrastive noise loss, illustrated in Fig \ref{fig:cont_noise_arch}.}
    \label{fig:pretext_arch}
\end{figure}

The architecture we designed for pre-training comprises three autoencoders, one for each type of omics data, and is shown in Fig \ref{fig:pretext_arch}. Our codebase also supports the usage of a common encoder and decoder for all three omic types, but since the inputs of the three omic types are different in size, we used some fully connected layers to downsample them to the same size and the rest of the encoder is shared. The pretext loss minimised during pre-training is a weighted sum of the losses described below. The codebase supports more SSL losses not described here, such as Maximum Mean Discrepancy (MMD) loss and latent reconstruction loss.

\subsubsection{Reconstruction loss}
Let's denote the input data $x$ of $i$th omic type as $x_i$ and the reconstructed data as $x'_i$. Let there be N number of omic types. In our case, the value of N is three. As given below, the reconstruction loss can be formulated as the mean squared error loss between input and reconstructed omics data.
\begin{equation}
\mathcal{L}_{\text {reconstruction}}=\frac{1}{N} \sum_{i=1}^{N} MSE\left(x_{i}, x_{i}^{\prime}\right)
\end{equation}

To make the network robust to noise, we performed partial corruption of one omic type and made the network recover the entire feature set, including the other omic types. For this purpose, we divided the almost 60,000-dimensional gene expression data into 23 subsets, each corresponding to the chromosome on which the gene is located. The same is done for around 400,000 dimensional DNA methylation data. The model is then trained to reconstruct the input data when some or all subsets of a specific omic type are masked. The masking methods used are zeroing out and adding Gaussian or swap noise. A random masking method is chosen during each epoch. For instance, 6, 12, 18 random subsets or all 23 subsets of gene expression data can be corrupted. The model can then be asked to reconstruct all the input features, including the corrupted gene expression features and DNA methylation and miRNA expression values. Here, a higher weightage is given in the loss function for recovering the corrupted omic type. Let $x_{1}$, $x_{2}$ and $x_{3}$ be gene expression, DNA methylation and miRNA expression features respectively. If we mask gene expression features, the reconstruction loss can be modified as
\begin{equation}
\mathcal{L}_{\text {reconstruction }}=\frac{1}{N} \left( 0.5 *  MSE\left(x_{1}, x_{1}^{\prime}\right) + 0.25 *  MSE\left(x_{2}, x_{2}^{\prime}\right) + 0.25 *  MSE\left(x_{3}, x_{3}^{\prime}\right) \right)
\end{equation}
Another novel pretext task that we designed is masked subset or chromosome prediction. As described above, gene expression and DNA methylation data were divided into subsets, and random subsets were masked in each epoch. We made the network predict which subsets were masked by feeding the representations from the encoder to a masked chromosome prediction module. 

\subsubsection{Contrastive alignment loss}
\label{subsec:cont_loss}
The latent representations $h_1$, $h_2$ and $h_3$ were passed through a projection network to obtain projections $z_1$, $z_2$ and $z_3$. 
The alignment loss introduced in CLIP (Contrastive Language-Image Pre-Training) \cite{clip} was used to compute alignment between the pairs ($z_1$, $z_2$), ($z_1$, $z_3$) and ($z_2$, $z_3$). The idea is that like text and image provide different types of information about a concept, various omic types contain different information about a patient's tumour. These multiple views are aligned using a contrastive loss.




\subsubsection{Contrastive noise loss}
\begin{figure}
    \centerline{
    \includegraphics[scale=0.37]{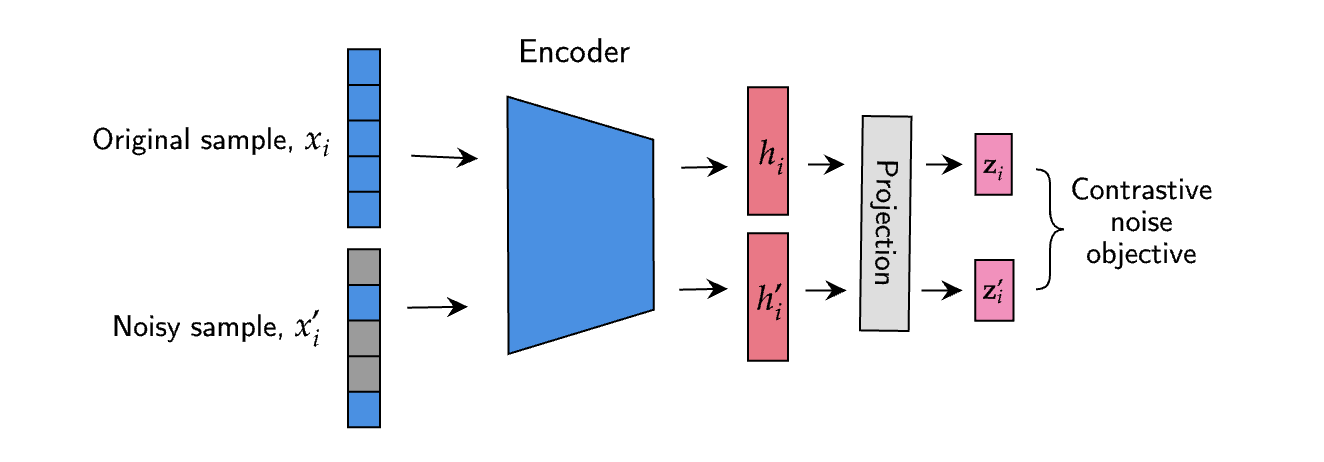}}
    \caption{Illustration of contrastive noise objective: Original and noisy samples are fed to the encoder one after the other to obtain their corresponding representations and projections between which the contrastive noise objective is calculated. This process is repeated for each omic type.}
    \label{fig:cont_noise_arch}
\end{figure}

Let the latent representation from $i$th omic type $x_i$ be denoted as $h_i$. By feeding noisy sample $x_i^{'}$ to the same encoder, we can produce $h_i^{'}$. By passing $h_i$ and $h_i^{'}$ through the projection layer, we obtain $z_i$ and $z_i^{'}$ respectively. Contrastive noise loss is computed between each pair $(z_i,z_i^{'})$. This is illustrated in Fig \ref{fig:cont_noise_arch}. The contrastive noise loss we implemented is the one introduced in the work Barlow Twins \cite{barlowtwins}. Our codebase also supports the usage of NT-Xent loss \cite{ntxent} and SimSiam loss \cite{simsiam} as both contrastive alignment and noise losses.

\subsubsection{Distance loss}
Using this loss, the distances between pairs of latent representations $(h_1, h_2)$, $(h_2,h_3)$ and $(h_1,h_3)$ are minimised. This ensures that representations from multiple omic types are consistent with each other. The distance loss can also be computed between the projections. Computing it between latent representations gave better results.
\begin{equation}
\mathcal{L}_{\text {distance}}=MSE(h_1, h_2) + MSE(h_2,h_3) + MSE(h_1,h_3)
\end{equation}



\subsection{Downstream task: Cancer type classification}
\begin{figure}
    \centerline{
    \includegraphics[scale=0.35]{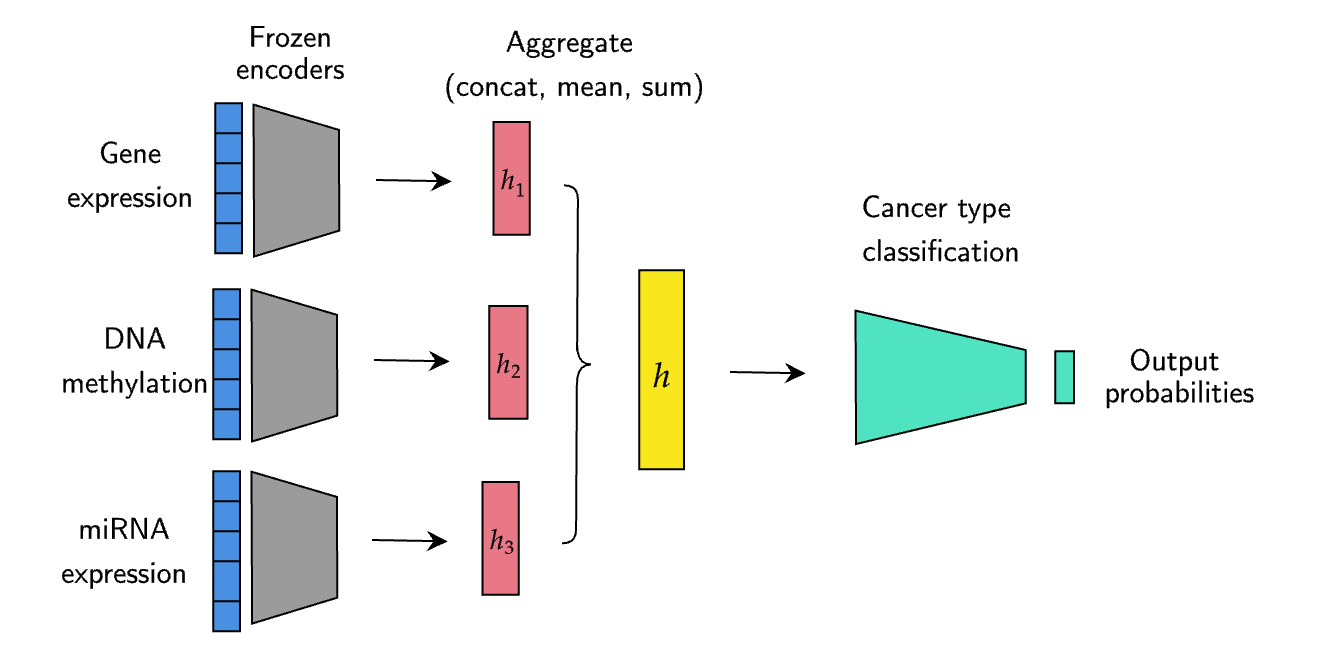}}
    \caption{Downstream module: The encoders are frozen after pre-training and representations from the encoders are aggregated by concatenating them, taking their mean or summing them up. This aggregated representation is then passed through the downstream network to predict the patient's tumour type.}
    \label{fig:ds_arch}
\end{figure}

Once the encoders and decoders are trained to minimise the pretext loss, the layers of encoders are frozen and attached to the downstream network to perform cancer type classification, as shown in Fig \ref{fig:ds_arch}. The dataset consists of 33 cancer types. Each patient's sample is a tissue that could be either normal or cancerous, belonging to one of these classes. The loss function for the downstream classification task is formulated as follows
\begin{equation}
    \mathcal{L}_{\text {classification}} =  CE(y, {y}')
    \label{eq:class}
\end{equation}
Here, $y$ is the label, ${y}'$ the prediction and CE the cross entropy loss.

\subsubsection{Handling missing omic types}
Our framework is suitable for handling missing omic types. For pre-training with missing omic types, encoders for only the available omic types can be trained. This is possible since we have separate encoders for each omic type. We can also train the downstream network using new samples that contain missing omic types. For instance, if gene expression is the missing omic type for a new sample from lung tumour, the pre-trained gene expression encoder can be used to generate gene expression representation for this new sample. This representation could be the average gene expression representation of all lung cancer samples. This way, information about the missing omic types can be generated in their absence. Alternatively, we could decide not to use the pre-trained encoder for missing omic types and aggregate representations (mean or sum aggregation) from available omic types. This flexibility allows the usage of datasets which contain different types of omics data for pre-training and downstream training. This is important as datasets usually differ in the type of omics data they contain.

\section{Experiments and results}
\subsection{Implementation}
Code and links to download the datasets are available at\\ \href{https://github.com/hashimsayed0/self-omics}{https://github.com/hashimsayed0/self-omics}. We used Pytorch Lightning \cite{torchlightning} to build the models.

\subsection{Semi-supervised learning}
\begin{figure}[!htbp]
    \centerline{
    \includegraphics[scale=0.7]{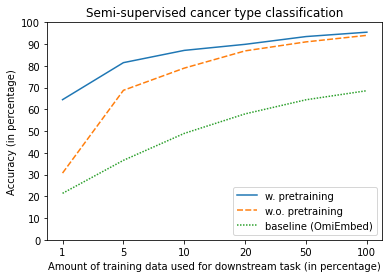}}
    \caption{Semi-supervised cancer type classification performance on multi-omics data: x-axis shows the percentage of training data used during downstream training and the y-axis denotes the accuracy. The encoders were kept frozen during downstream training.}
    \label{fig:semi_sup_graph}
\end{figure}

To evaluate the effect of pre-training, we trained the model in semi-supervised fashion. The model was first pre-trained on the entire training set and was then trained for the downstream task using only part of the training set. The encoders were also kept frozen and not allowed to be optimised for the downstream task. Fig \ref{fig:semi_sup_graph} shows the leap in performance provided by our pre-training approach over training the downstream network with random initialisation and OmiEmbed \cite{omiembed} which is the state-of-the-art approach in cancer type classification using multi-omics data. A performance comparison between the methods based on metrics is given in Table \ref{tab:semi_sup}. 

\begin{table}[]
\tbl{Performance metrics of cancer type classification using 1\% training data during downstream training.}
{
\begin{tabular}{@{}ccccccc@{}}
\toprule
\multirow{2}{*}{\textbf{Method}} & \multirow{2}{*}{\textbf{Omic type(s)}} & \multicolumn{5}{c}{\textbf{1\% training data}}                                             \\ \cmidrule(l){3-7} 
                                 &                                        & \textbf{Accuracy} & \textbf{F1}    & \textbf{AUC}   & \textbf{Precision} & \textbf{Recall} \\ \midrule
OmiEmbed                         & multi-omics (A,B,C)                    & 21.37             & 7.82           & 73.58          & 6.77               & 14.74           \\ \midrule
w.o. pretraining                 & multi-omics (A,B,C)                    & 30.69             & 19.1           & 73.03          & 32.2               & 22.85           \\ \midrule
\multirow{4}{*}{w. pretraining}  & gene exp. (A)                          & 13.9              & 4.21           & 57.99          & 3.71               & 9.67            \\
                                 & DNA meth. (B)                          & 32.98             & 20.47          & 70.59          & 21.45              & 23.66           \\
                                 & miRNA exp. (C)                         & 42.75             & 27.21          & 82.51          & 28.92              & 32.2            \\
                                 & multi-omics (A,B,C)                    & \textbf{64.45}    & \textbf{43.33} & \textbf{82.95} & \textbf{43.99}     & \textbf{49.83}  \\ \bottomrule
\end{tabular}
}
\label{tab:semi_sup}
\end{table}

\subsection{Ablation study}
We ran experiments to analyse the effects of removing various components of the pretext loss one at a time. This usually helps identify the essential components of the pretext loss and evaluate the method's robustness. Fig \ref{fig:ablation_latagg} shows the effect on downstream performance due to the removal of various components of pre-training loss. The performance is robust to such removals and is not overly dependent on any component.


\subsection{Latent aggregation method}
\begin{figure}[!htbp]
    \centerline{
    \includegraphics[scale=0.535]{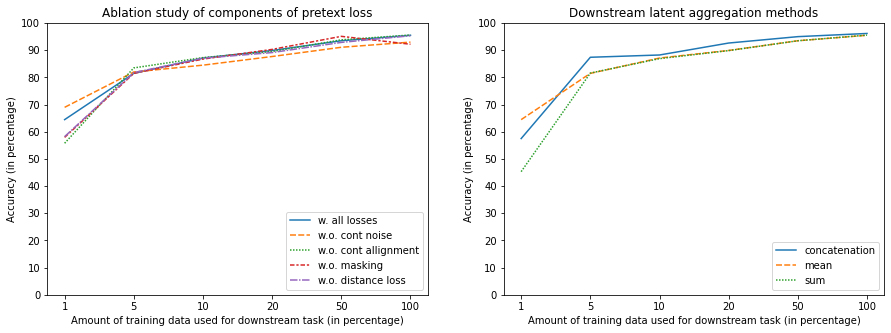}}
    \caption{The plot on the left shows the effect of removing various components of pretext loss, and the one on the right shows the performance variation with different downstream latent aggregation methods. The x-axis shows the percentage of training data used during downstream training, and the y-axis denotes accuracy.}
    \label{fig:ablation_latagg}
\end{figure}

As we have separate encoders for each omic type, representations from the encoders have to be aggregated to be passed to the downstream network. We experimented with various aggregation methods, including mean, concatenation and sum. Fig \ref{fig:ablation_latagg} shows that concatenation performs slightly better than other methods. 


\subsection{t-SNE Visualisation}
\begin{figure}
    \centerline{
    \includegraphics[scale=0.60]{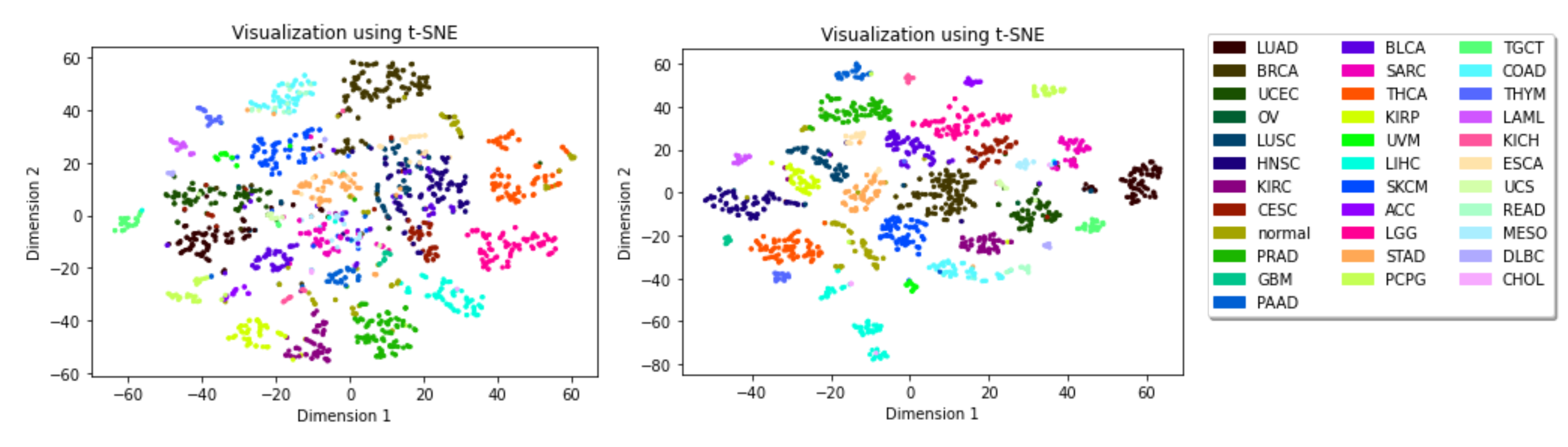}}
    \caption{t-SNE Visualisation: The plot on the left is produced using a model trained on one per cent of training data for downstream training, and the one on the right refers to the model that used the entire training set for downstream training. The legend shows the TCGA codes for the cancer types that the colours represent.}
    \label{fig:tsne}
\end{figure}

To visualise the model's discriminative ability, we fed the latent representations produced by a trained model to t-SNE \cite{tsne}. Fig \ref{fig:tsne} shows how the model clusters test samples from the same cancer type together. It is interesting to note how well the model is able to cluster cancer types even using 1\% training data for the downstream task.

\section{Discussion}
By analysing the results of the experiments, it is clear that our approach works well with less training data, thanks to efficient pre-training. Although OmiEmbed performs well with an unfrozen encoder when the whole training set is provided during downstream training, it fails to achieve decent performance when encoder layers are frozen and a limited amount of training data is used. The performance of our approach with frozen encoders is comparable to the performance of OmiEmbed with unfrozen encoder, as reported in their paper \cite{omiembed}. With ablation studies, we show how the model is not entirely dependent on any particular component of the pretext task. The contrastive noise objective helps the encoders become robust to noise and be able to identify samples from their distorted versions. The contrastive alignment objective makes the encoders learn similar and discriminative information from representations of different omic types of the same patient. Reconstructing the full feature set from the representation of one omic type forces an encoder to learn information about the two other omic types from this omic type. By asking the network to recover masked gene expression data, we make it rely on DNA methylation and miRNA expression data. From our experiments, we found out that masking the latter two omic types did not improve performance. This is also in line with our understanding that gene expression is influenced by DNA methylation and miRNA expression. While there is a significant difference in performance between different experiment settings when less than 10\% of training data is used, the performance converges as a higher amount of data is used. 

\section{Conclusion}
We began by discussing various SSL approaches used in tabular domain and methods that integrate multi-omics data. After describing the dataset, we formulated the various components of our pretext task. The main idea behind using these components was to make the encoders learn what is common and specific about different omic types based on patients' profiles. This would help the encoders produce relevant features for the downstream tasks. To evaluate our approach, we designed a semi-supervised framework and ran experiments. We showed that in this framework, our approach outperforms the state-of-the-art method. We performed ablation studies to analyse our approach and its robustness. We then discussed key insights from the results and explained our findings. This work has shown that pre-training with a huge dataset like TCGA with efficient components improves downstream performance in various settings. Our approach also offers the flexibility to use different datasets for pre-training and downstream training and is suitable for handling missing omic types. A limitation of this approach is that the features present in the pre-training dataset need to be available in the downstream dataset to perform pre-training and downstream training on different datasets. Another limitation is that the models trained in this framework contain many parameters and require a good amount of CPU and GPU memory to load the dataset and train the model. This work can be further extended to perform zero-shot classification of rare cancer types. To do this, we need to develop a model that learns about rare cancers from common cancers. This might require representing cancer types like words in latent spaces\cite{ssl_nlp1, ssl_nlp2}. It could be useful to investigate models like gene2vec \cite{gene2vec} for this purpose.

\bibliographystyle{ws-procs11x85}
\bibliography{ws-pro-sample}

\end{document}